%% file: egpaper.tex
\documentclass[10pt,twocolumn,letterpaper]{article}

\usepackage{wacv}
\usepackage{times}
\usepackage{epsfig}
\usepackage{graphicx}
\usepackage{amsmath}
\usepackage{amssymb}
\usepackage{booktabs}
\usepackage{tikz}
\usepackage{comment}
\usepackage{amsfonts}
\usepackage{bbold}
\usepackage{color}
\usepackage{multirow}
\usepackage{gensymb}
\usepackage{pgfplots}
\usepackage{caption}
\usepackage{arydshln}
\usepackage{dblfloatfix}
\usepackage[accsupp]{axessibility}

\def\wacvPaperID{597} 

\wacvalgorithmstrack   

\wacvfinalcopy 

\usepackage{hyperref}

\begin{document}

\title{ImPosing: Implicit Pose Encoding for Efficient Visual Localization}

\author{Arthur Moreau\\
MINES ParisTech\\
Huawei Technologies\\
\and
Thomas Gilles\\
MINES ParisTech\\
Huawei Technologies\\
\and
Nathan Piasco\\
Huawei Technologies\\
\and
Dzmitry Tsishkou\\
Huawei Technologies\\
\and
Bogdan Stanciulescu\\
MINES ParisTech\\
\and
Arnaud de La Fortelle\\
MINES ParisTech\\
}

\maketitle
\thispagestyle{empty}

\begin{abstract}
   We propose a novel learning-based formulation for visual localization of vehicles that can operate in real-time in city-scale environments. Visual localization algorithms determine the position and orientation from which an image has been captured, using a set of geo-referenced images or a 3D scene representation. Our new localization paradigm, named Implicit Pose Encoding (ImPosing), embeds images and camera poses into a common latent representation with 2 separate neural networks, such that we can compute a similarity score for each image-pose pair. By evaluating candidates through the latent space in a hierarchical manner, the camera position and orientation are not directly regressed but incrementally refined. Very large environments force competitors to store gigabytes of map data, whereas our method is very compact independently of the reference database size. In this paper, we describe how to effectively optimize our learned modules, how to combine them to achieve real-time localization, and demonstrate results on diverse large scale scenarios that significantly outperform prior work in accuracy and computational efficiency.
\end{abstract}

\section{Introduction}

\input{sections/introduction}

\section{Related work}

\input{sections/related_work}

\section{Method}

\input{sections/method}

\section{Experiments}

\input{sections/experiments}

\section{Discussion}

\input{sections/discussion}

\section{Conclusion}

\input{sections/conclusion}

\clearpage
{\small
\bibliographystyle{ieee_fullname}
\bibliography{egbib}
}

\clearpage
\appendix
\begin{center}
\textbf{\large Supplementary Materials}
\end{center}
\vspace{0.5cm}

This document presents further analysis on our method. We present additional ablation studies, latent space visualization, results of the attached video and reproducibility details. We invite readers to view the supplementary video where localization results are shown on a wide range of scenarios.

\section{Ablation study on the pose encoder capacity}

All experiments in the main paper report results with 4 layers in the pose encoder MLP network. We evaluate the localization results with different pose encoder capacity on Neighborhood and Countryside scenes from the 4seasons dataset\cite{wenzel2020fourseasons}.

\input{tables/pose_encoder_ablation}

Surprisingly, we observe that MLPs with a single hidden layer perform better on both scenes. The reason is not very clear: more capacity should not degrade performance except in case of overfitting, which is not the case here because the training loss is lower for smaller models as well. It might be that bigger MLPs just take more time to converge, we stopped the experiment after 250 epochs.

\section{Ablation study on the similarity score}

We tried to alternatives to cosine similarity for computing the score between image and camera pose latent vectors. A first alternative is based on L2 distance between the image and map signatures:

\begin{equation}
    s(I,p) = 1 - \lVert f_{I}(I) - f_{M}(p) \rVert_{2} \quad \mathbb{1}_{1 - \lVert f_{I}(I) - f_{M}(p) \rVert_{2} > 0}
\end{equation}

Then, we also tried to learn this step with a 2 layers MLP, which takes $f_{I}(I) - f_{M}(p)$ as input, uses a ReLU activation in the hidden layer and outputs a score through the sigmoid activation. 
These solutions are compared on the Neighborhood scene from the 4seasons dataset\cite{wenzel2020fourseasons} on figure~\ref{fig:match_mod}. These scores are supervised with the target scores described in section 3.2 on the main paper.

\input{tables/matching_module_ablation}

The ablation confirms that cosine similarity performs better than other alternatives to compute the score between image-pose pairs.

\section{Supplementary video}

The attached video file shows sequential qualitative results of single scenes ImPosing models (corresponding to quantitative results of tables 1 and 2 in the main paper) on Daoxiang Lake~\cite{DA4AD_2020_ECCV} and 4Seasons datasets~\cite{wenzel2020fourseasons}.

The input image is displayed on the top left corner. The right part shows the current predicted trajectory in red, ground truth poses in green and training trajectories in gray. The bottom left corner displays the 256 best candidates selected for pose averaging in red, the predicted pose in black and the groundtruth pose in green. Finally, the last plot shows the score of all candidates in the entire map from transparent ($s=0$) to red ($s=1$).

Scenes are displayed in the following order : Daoxiang Lake (00:00 to 01:03), Neighborhood (01:04 to 02:09), Office Loop (02:10 to 02:40), Business campus (02:41 to 03:38), City Loop (03:39 to 03:59), Countryside (04:00 to 04:44), Old Town (04:45 to 05:00).

\begin{figure}[h]
\centering
\begin{minipage}{0.22\textwidth}
\includegraphics[width=\linewidth]{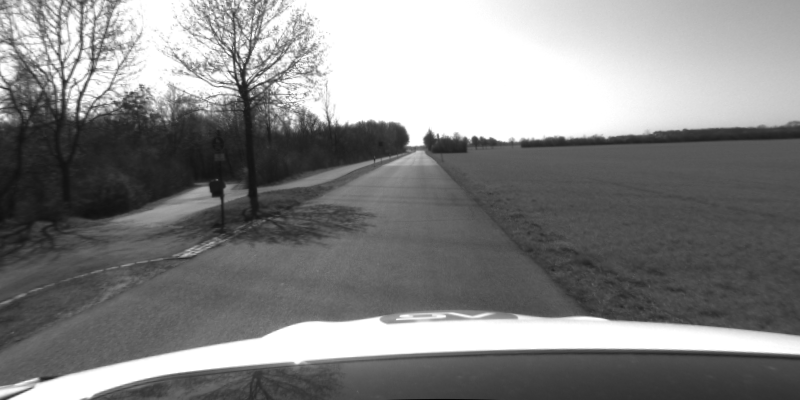}
\end{minipage}
\begin{minipage}{0.22\textwidth}
\includegraphics[width=\linewidth]{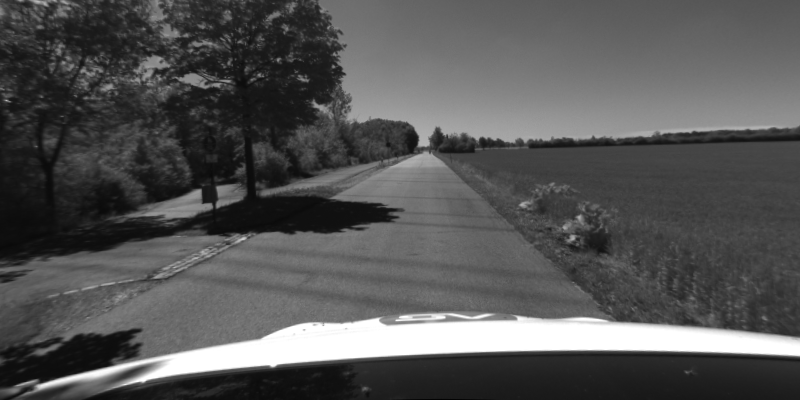}
\end{minipage}
\begin{minipage}{0.22\textwidth}
\includegraphics[width=\linewidth]{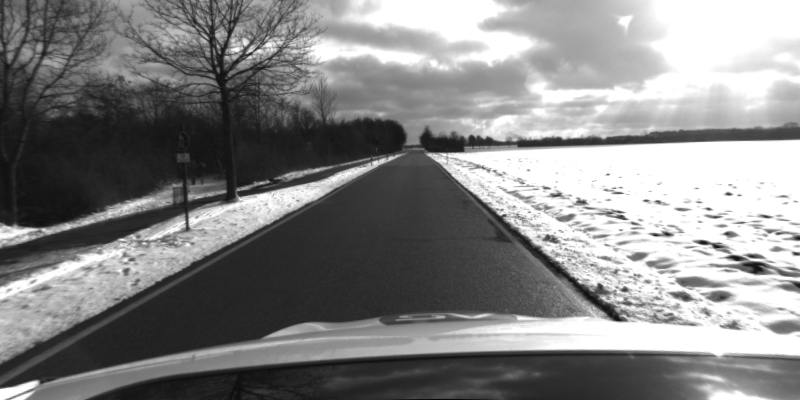}
\end{minipage}
\hspace{2pt}
\begin{minipage}{0.22\textwidth}
\includegraphics[width=\linewidth]{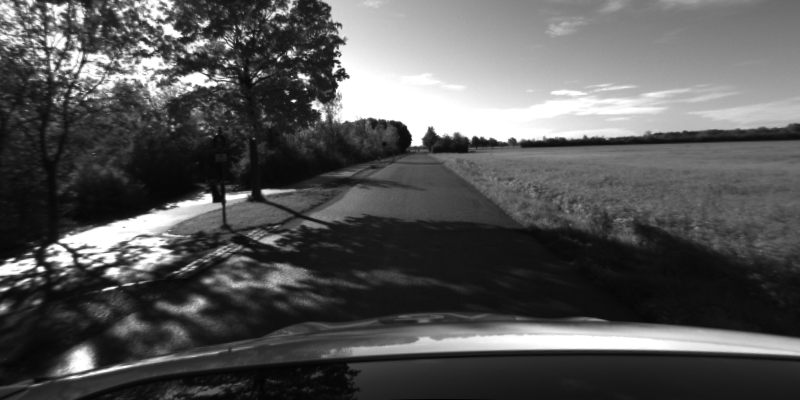}
\end{minipage}
\caption{\label{fig:countryside} \textbf{Featureless environments and varying weather conditions.} Test is performed on the image on the right, while the network has been trained with 3 recordings with different lightning conditions. Our method is able to provide a coarse localization in these scenarios, where as image retrieval and pose regression competitors fail.}
\end{figure}

\begin{figure}[h!]
\centering
\begin{minipage}{0.33\textwidth}
\includegraphics[width=\linewidth]{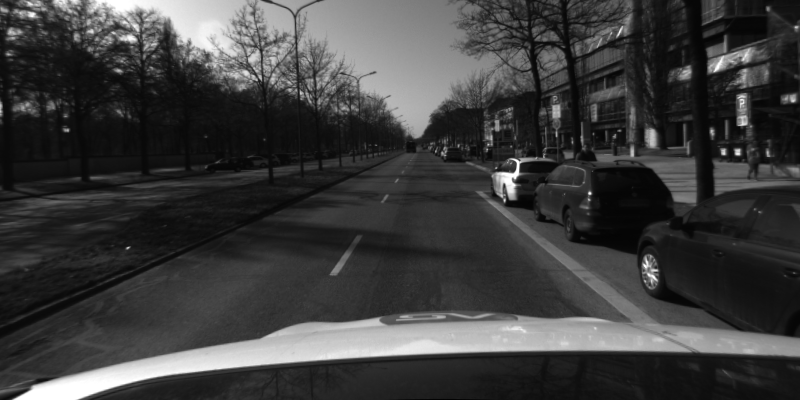}
\end{minipage}
\begin{minipage}{0.25\textwidth}
\includegraphics[width=\linewidth]{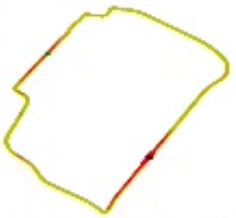}
\end{minipage}
\begin{minipage}{0.33\textwidth}
\includegraphics[width=\linewidth]{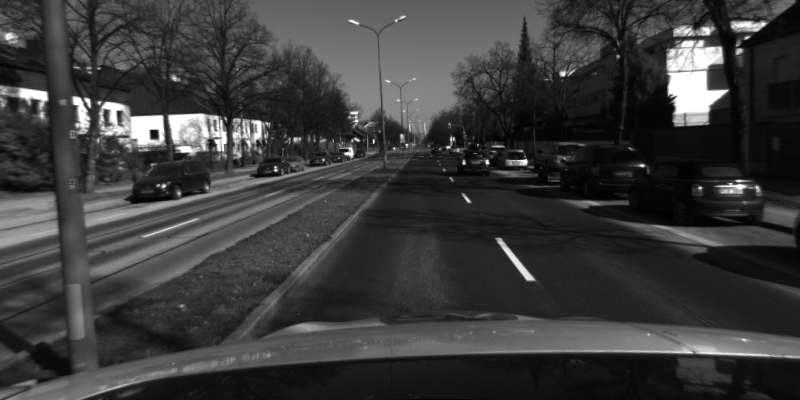}
\end{minipage}
\caption{\label{fig:ambiguity} \textbf{Multimodal score distribution in ambiguous cases.} Many road areas present similar structure and appearance, introducing ambiguities in the localization task. In this scenario from the City Loop scene, the model outputs high scores for areas depicted in left and right images, which are very far one from each other. By refining the estimate in further steps, the model is able to solve this ambiguity in most cases.}
\end{figure}

These video samples show clearly advantages and limitations of our method:
\begin{itemize}
    \item Coarse localization is correct most of the time, even in large maps with repetitive and featureless environments (see figure~\ref{fig:countryside}).
    \item In ambiguous scenarios, our method provides a multimodal distribution of scores in first iterations and then solves the ambiguity in further steps (see figure~\ref{fig:ambiguity}).
    \item Sequences of predictions are not temporally smooth, because each frame is treated independently in this experiment. In practice, this can be solved by filtering with a motion model, similar to~\cite{coordinet}.
    \item Precise pose estimation is sometimes inaccurate but sufficient to provide a lane level localization for navigation of autonomous vehicles.
\end{itemize}

It should also be noted that experiments on 4seasons dataset are extreme scenarios where the quantity of available data is small w.r.t. to the challenges introduced by weather conditions.

\section{Latent space visualization}

\begin{figure*}[h]
   \centering
   \includegraphics[width=0.7\linewidth]{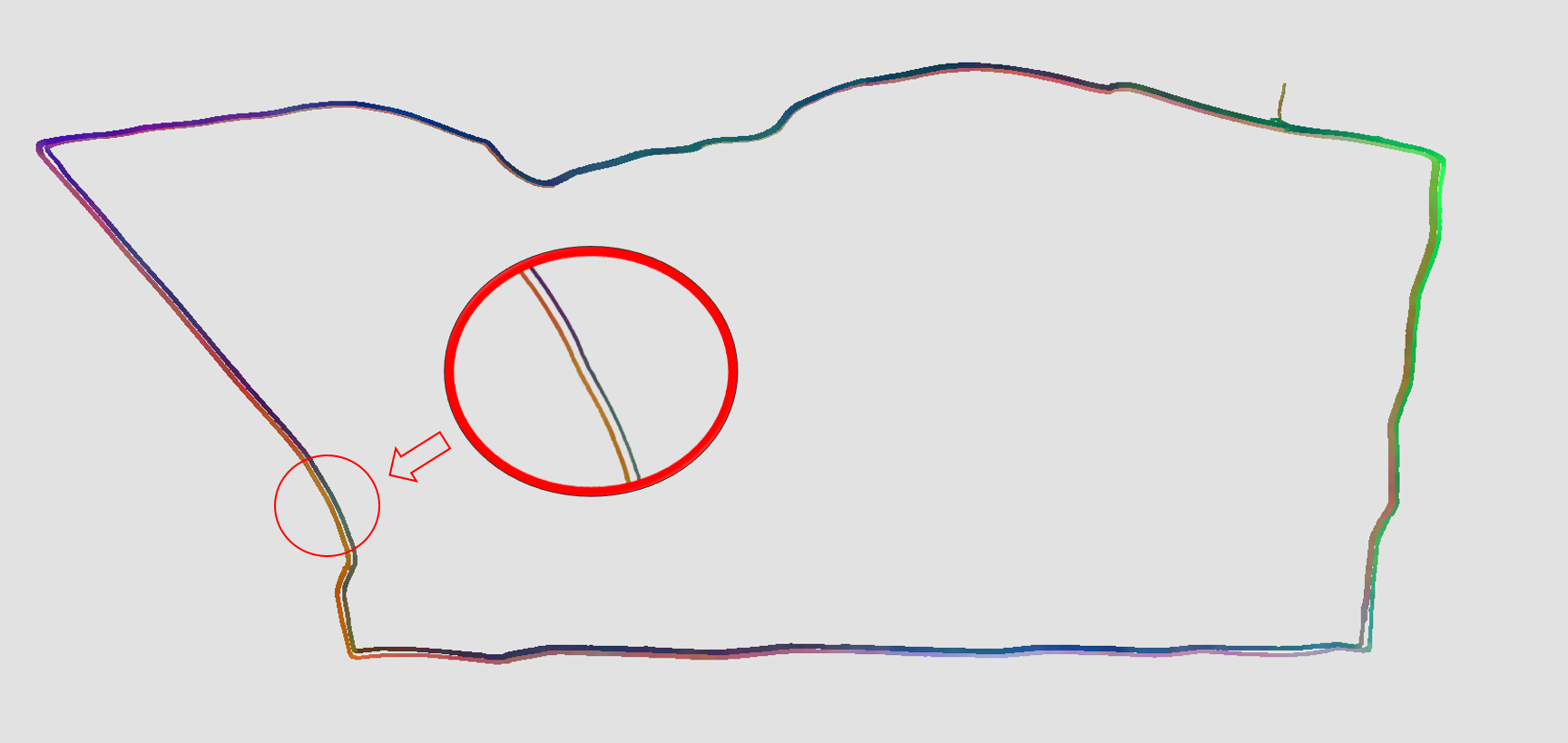}
   \caption{\textbf{Latent space visualization.} Training poses, colored by the 3 principal components of map descriptors. Poses with similar colors are close in the latent space. Opposite ways of the same road are represented by dissimilar representations. Best viewed in color}
   \label{fig:map_visu}
\end{figure*}

We attempt to visualize the structure of the latent space learned by ImPosing. We compute the latent vector of all reference poses of the Daoxiang Lake map. Then we compute a PCA of the 256 dimensional vectors and display it on the map in figure~\ref{fig:map_visu}. We observe that our pose encoder learns a smooth representation of the map, where close representations share similar visual content.

\section{Datasets preparation}

This section contains dataset splits used in our experiments to ensure reproducibility. To the best of our knowledge, 4Seasons~\cite{wenzel2020fourseasons} and Daoxiang Lake datasets~\cite{DA4AD_2020_ECCV} had not been used previously to evaluate direct learning-based methods. After preparing all the absolute poses of a map, we normalize the positions between -0.5 and 0.5, such that the networks converge faster in kilometers scale maps.

\subsection{Oxford RobotCar dataset}

The dataset can be downloaded \href{https://robotcar-dataset.robots.ox.ac.uk/}{here}. We replicate experiments from previous methods using undistorted front camera images:

\input{tables/oxford_split}

\subsection{Daoxiang Lake dataset}

The dataset can be downloaded \href{https://apollo.auto/daoxianglake.html}{here}. Vehicles are equipped with multiple sensors but we only use the front cameras images with associated vehicle poses. We don't use the train/test split provided by the dataset because the test set is not an entire held out sequence (images from the same sequence has been observed during training) and then is not a realistic test scenario.

\input{tables/daoxianglake_split}

\subsection{4seasons dataset}

The dataset can be downloaded \href{https://vision.cs.tum.edu/data/datasets/4seasons-dataset}{here}. Absolute poses are generated using available \href{https://github.com/Artisense-ai/libartipy}{Python tools}. We use keyframes from the left camera only.

\input{tables/4seasons_split}

\end{document}

%% file: sections/introduction.tex
Positioning systems are a necessary component for automated vehicles, mobile robots and augmented reality applications. The precise ego-position inside of a known environment can be recovered in multiple ways using a wide range of sensors. Visual-based localization algorithms~\cite{piasco2018survey} predict the 6 degrees of freedom camera pose of a query image, given a set of reference images captured in the environment and labeled with corresponding poses. 

We aim to develop relocalization algorithms able to operate efficiently in embedded devices of autonomous vehicles in a deployment scenario where the target area is wide and collected datasets are large. This problem is challenging due to kilometer-scale maps and dynamic outdoor environments. Most accurate visual localization methods~\cite{sarlin2021pixloc,DA4AD_2020_ECCV} first retrieve a coarse localization (i.e. which area is depicted in the image) before computing an accurate camera pose with geometric reasoning by connecting 2D image features to 3D points stored in memory with their corresponding descriptors. The resulting accuracy comes at the cost of a high memory footprint and low latency which increase with the environment size and the reference database. Direct learning-based methods~\cite{PoseNet,coordinet,brachmann2019esac} circumvent this limitation by learning the entire task with a single neural network that directly regresses the camera pose from the image. This solution is convenient for embedded deployment :  high throughput, low memory footprint and ability to benefit from large amount of data during training without sacrificing test time efficiency. On the other hand, image features extraction and map memorization are entangled in the network's weights, resulting in a limited accuracy~\cite{Sattler2019}, slow scene specific training and poor ability to adapt to large environments~\cite{brachmann2019esac}. Our proposal improves the accuracy and the scalability of direct learning-based methods while preserving the computational efficiency properties.

The common approach to represent scenes in computer vision is to use explicit representations such as point clouds, octrees, voxels or meshes. However, all of them store discrete information, while the underlying signal they represent is inherently continuous. As a consequence, these representations involve a trade-off between resolution and memory consumption. Recently, implicit neural representations~\cite{xie2021neuralfield}, that connect scene coordinates to latent codes with a neural network, have shown great success for many computer vision tasks thanks to their ability to model continuous signals embedded into compact network's weights~\cite{nerf}.

In this paper, we propose a new direct approach for visual localization in large scenes that perform better than pose regression methods by dissociating image and map encodings, while avoiding the computational cost and memory footprint of structure-based methods thanks to an implicit map representation. The core idea is to connect image and camera pose representations, which are learned separately by two distinct neural networks, in a common latent space. We use an implicit neural representation to encode a specific viewpoint in the scene (i.e. a 6-DoF camera pose) into a higher dimensional vector. With this formulation, the continuous representation of any camera pose in the scene (even a pose not observed in reference images) can be computed in a single network forward pass. We take advantage of this property to solve the localization task by searching the poses candidates which are the most similar to the learned image representation. To do so, we introduce a hierarchical sampling process able to retrieve the correct camera viewpoint using only a few batched queries on the pose encoder network. Our localization method, called Implicit Pose Encoding (ImPosing), provides real-time sub-metric localization performances that can be rapidly deployed on large areas.

We evaluate our system on a wide range of visual localization datasets, including several kilometers-scale road environments with challenging conditions (seasonal and appearance changes, limited training data). We observe that our method outperforms its regression-based competitors in terms of accuracy and training efficiency, especially in large-scale scenarios.

%% file: sections/related_work.tex
\label{sec:related_work}

\paragraph{Image-based localization.} Camera localization from RGB images for real-time application can be tackled by different classes of prior methods discussed below:

 \textbf{Absolute pose regression} addresses the problem through end-to-end supervised regression between the input image and the camera pose using  deep neural networks. PoseNet~\cite{PoseNet} is the pioneering work, and uses an encoder-decoder architecture where the encoder is a CNN pretrained on ImageNet and the decoder regresses the pose with fully connected layers. Since then, many architectural improvements have been proposed: notably, VidLoc~\cite{vidloc} incorporates spatio-temporal constraints using consecutive video frames,  AtLoc~\cite{AtLoc} uses an attention-based module before the decoding step, Xue et al.~\cite{graph_neural_network} model the problem with graph neural networks, TransPoseNet~\cite{TransPoseNet} with transformers, and CoordiNet~\cite{coordinet} uses a fully-convolutional architecture with geometrical inductive biases in the decoder layers. The main advantages of this class of methods are the compatibility with real-time deployment thanks to fast inference, low memory requirements and uncertainty estimation~\cite{bayesianposenet,coordinet} which enables to filter out failure cases. The localization accuracy exhibited by absolute pose regression is limited compared to other methods~\cite{Sattler2019}, but has been observed to be highly dependent on the quantity and diversity of available training images, which can be improved with novel view synthesis~\cite{lens}. ImPosing does not not explicitly regress the pose of the camera but learns a latent representation which connects the query image to an implicit map. In the following, we show through experiments that this formulation is better suited than absolute pose regression for localization in large urban area.
  
  \begin{figure*}[t]
   \centering
   \includegraphics[width=0.8\linewidth]{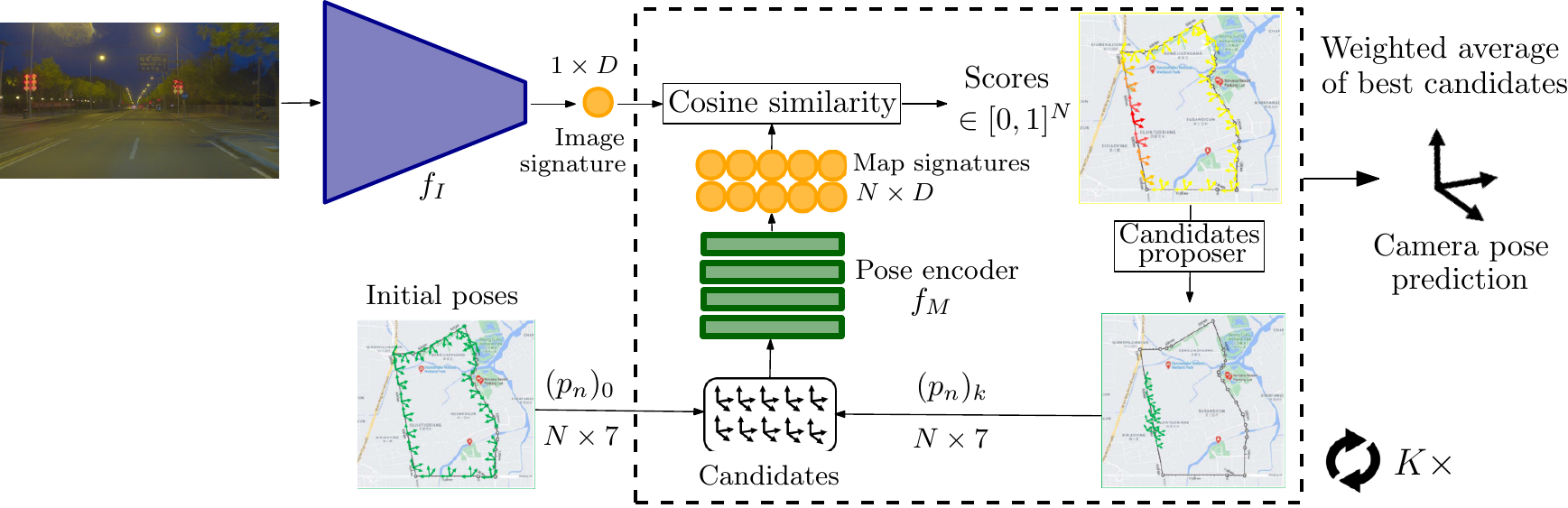}
   \caption{\textbf{Implicit pose encoding for hierarchical image localization.} A set of initial map signatures is compared to the image signature to determine the most probable localization of the camera. The similarity scores guides the selection of a new batch of pose candidates that are used to compute the new map signatures for the second refined localization step. This process is repeated multiple time to predict the final camera pose.}
   \label{msnet_pipeline}
\end{figure*}
  
\textbf{Scene coordinate regression} learns the correspondence between the 2D image features and 3D scenes coordinates of observable image patches. It enables to retrieve the camera pose using projective geometry, by solving the Perspective-N-Points problem robustly with RANSAC~\cite{RANSAC}. Seminal work on scene coordinate regression rely on RGB-D images and use random forest to store the 3D coordinates~\cite{forest_SCR}. Since then, the scene coordinate regression pipeline has been adapted to RGB images processed by fully convolutional networks~\cite{scr_nn,Brachmann_2018_CVPR}. The RANSAC step has been replaced by its differentiable counterpart DSAC~\cite{brachmann2017dsac}, and ESAC~\cite{brachmann2019esac} uses mixtures of expert to improve scaling to large environments. 
This class of methods exhibit higher accuracy than absolute pose regression and the efficiency enables real-time computation, however these methods are limited to relatively small environments~\cite{dsac*}. By considering global image description instead of local features extraction, ImPosing is able to scale up to larger scenes at the cost of minor loss in localization performances.
  
    \textbf{Image retrieval algorithms for localization} solve a slightly different task: instead of computing a pose for the query image, these methods retrieve the closest geo-referenced image from the query within a large database~\cite{netvlad,GARL17,RARS19,piasco_2019}. The top ranked images are used to define a coarse localization of the query image. Poses averaging~\cite{torii2011visual} or specific re-ranking based on GPS information~\cite{sattler2016large} are used to improve the localization accuracy. Image retrieval methods use global image descriptors obtained by features maps pooling~\cite{netvlad} or dense local features extraction~\cite{torii201524} to represent the discriminating content of the image. Nearest neighbour search in the descriptor space associates the query to the most similar examples in the database. These methods naturally scale to very large environments~\cite{cityscalelocation} but their accuracy is bounded  by the density and diversity of reference images in the scene. Such a large database is difficult to collect and enlarging it linearly increases the memory footprint and the nearest neighbour search computational cost. This property make image retrieval an appealing solution for visual place recognition but not convenient for camera pose estimation. Our method share similarities with image retrieval: a global image descriptor is matched against the map. In our case, the geo-referenced image batabase is replaced by an implicit map representation. As a result, we can compute the descriptor of any camera pose in the map instead of being limited a finite set of reference images. Morever, for a given scene, larger datasets improve the resolution of the map representation without increasing the memory footprint of our map, stored as network weights.
 
\textbf{Structure-based methods} compare local 2D image features to a 3D model to estimate the camera pose. 
2D features are extracted from the query image using a CNN such as SuperPoint~\cite{detone18superpoint}, and matched against the 3D model~\cite{sarlin20superglue} to establish robust 2D-3D correspondences, that enable to compute the pose with PNP + RANSAC~\cite{sarlin2019coarse} or by Levenberg-Marquadt optimization \cite{von2020gn,sarlin2021pixloc}.
The 3D model, usually represented as a point cloud of descriptors, enables to use geometric reasoning to solve the task. However, in large dynamic environments, highly accurate 3D reconstructions are challenging to make and memory demanding. Relative pose regression~\cite{rpr_laskar,relocnet,camnet} from nearest images can alternatively be used to predict the pose, but the storage requirement is even bigger. ImPosing does not rely on a 3D model of the scene and operate only with images and references poses.

\paragraph{Implicit representations.} Neural networks performances highly depend on the representation used for a given space. Recent research has shown that using fully-connected neural networks to represent 3D data offers many benefits: the representation is continuous, memory-efficient and convenient to learn in any differentiable pipeline~\cite{xie2021neuralfield}. Successful examples of neural representations include 3D shapes~\cite{Park_2019_CVPR,Atzmon_2020_CVPR}, sound~\cite{sitzmann2019siren}, neural rendering of static~\cite{nerf,sitzmann2019srns,Local_Implicit_Grid_CVPR20} and dynamic scenes~\cite{nerfw} or real-time RGB-D SLAM~\cite{Sucar:etal:ICCV2021}.

In this paper, we aim to learn an effective representation of the map for camera relocalization inside of a given scene. The map is given as a set of images with 6D camera poses: a 3D translation vector and a 3D rotation represented by quaternions, euler angles, axis-angle or rotation matrix. Zhou et al.~\cite{Zhou_2019_CVPR} have demonstrated that none of these rotation representations are continuous, in the sense of continuously mapping coordinates to a latent space produced by a neural network, which is precisely our problem of interest. Zhu et al.~\cite{zhu_cvpr} proposed a learned camera pose representation which is beneficial for view synthesis and pose regression. We propose to use a related camera pose representation optimized to be directly matched against the input image representation, enabling pose estimation by iterative sampling and evaluation of pose candidates.

%% file: sections/method.tex
\label{sec:method}

Our method, ImPosing, estimates the 6-DoF camera pose $(t,q) \in SE(3)$ of a query image $I$, where $t$ is a translation vector and $q$ is a unit quaternion. We train our solution using a reference dataset of posed images $(I_{k})$ collected in the target area and we do not make use of an additional 3D model of the scene. 

The proposed algorithm, presented in figure~\ref{msnet_pipeline}, computes a vector that represents the image through the image encoder. Then, the camera pose is searched by evaluating initial pose candidates distributed across the map. Poses are processed by the pose encoder to produce a latent representation that can be matched against the image vector. Each pose candidate receives a score, based on distance to camera pose. High scores provide a coarse localization prior which is used to select new candidates. By repeating this process several times, our pool of candidates converges to the actual camera pose.


\subsection{ImPosing localization process}
\label{method_1}

This section describes the localization process step by step from the image to the final camera pose estimate, displayed in figure~\ref{msnet_pipeline},

\paragraph{1. Image encoder:} we compute a global image features vector $f_{I}(I)\in \mathbb{R}^{d}$ from the input query $I$ using our image encoder. The encoder architecture consists in a pretrained CNN backbone followed by a Global Average Pooling, and a fully-connected layer with $d$ output neurons. The feature vector is one order of magnitude smaller than global image descriptors commonly used in image retrieval (we use $d=256$ whereas Revaud et al.~\cite{RARS19} use $d=2048$) in order to efficiently compare it to a large set of pose candidates at later steps.

\paragraph{2. Initial pose candidates:} Our starting point is a set of $N$ camera poses $(p_{n})_{0}$, sampled from the set of reference poses (= training poses). Through this initial selection, we introduce a prior for the localization process, similar to the anchors poses in~\cite{saha2018improved} or regression methods that compute relative instead of absolute pose~\cite{ding2019camnet}. We observed that the algorithm is robust to this choice: a 2D grid on the map yield similar results. 

\paragraph{3. Pose encoder:} Pose candidates are processed by a neural network which outputs latent vectors. This implicit representation learns the correspondence between camera viewpoints in a given scene and features vectors provided by the image encoder. First, following Tancik et al.~\cite{tancik2020fourier}, each component of the camera pose $(tx, ty, tz, qx, qy, qz, qw)$ is projected to higher dimension using Fourier features : $x \rightarrow (x, sin(2kx), cos(2kx))_{0 \leq k \leq 10}$, as it helps networks with low dimensional input to fit high frequency functions. Then, we use a MLP $f_{M}$ with 4 layers of 256 neurons and ReLU activations on hidden layers. Each set of pose candidates is computed in a single batched forward pass.

\paragraph{4. Similarity scores:} we obtain a similarity score $s$ by computing the cosine similarity between $f_{I}(I)$ and $f_{M}(p)$ for each image-pose pair $(I,p)$. We add a ReLU layer after the dot product, such that $s \in [0,1]$. Intuitively, we aim to learn high scores for poses candidates close to the actual camera pose. With this formulation, we can evaluate hypotheses on the camera pose and search for pose candidates with high scores. Formally, our score is defined by:

\begin{equation}
    s(I,p) = \frac{\langle f_{I}(I) , f_{M}(p) \rangle}{\| f_{I}(I) \| \|  f_{M}(p) \|} \mathbb{1}_{\langle f_{I}(I) , f_{M}(p) \rangle > 0}
\end{equation}

\paragraph{5. Candidates proposer:} new poses $(p_{n})_{k}$ are selected for the $k^{th}$ iteration based on scores obtained with poses $(p_{n})_{k-1}$ at the previous iteration. First, we select the poses with top $B=100$ higher scores $(h_{i})_{0 \leq i < B} \subset (p_{n})_{k-1}$. Then, new candidates are sampled from $(h_{i})$ in a Gaussian Mixture Model with density:
\begin{equation}
P(x) = \sum_{i=1}^{100} \pi_{i} \mathcal{N}(x | h_{i},\,v/k) \hspace{2mm} \text{where} \hspace{2mm} \pi_{i} = \frac{s(I,h_{i})}{\sum_{l=1}^{100} s(I,h_{l})}.
\end{equation}

$v=[v_{tx},v_{ty},v_{tz},v_{rx},v_{ry},v_{rz}]$ is the variance of the sampling process, a hyperparameter composed of a translation vector and Euler angles.

\begin{figure}[h]
   \includegraphics[width=\linewidth]{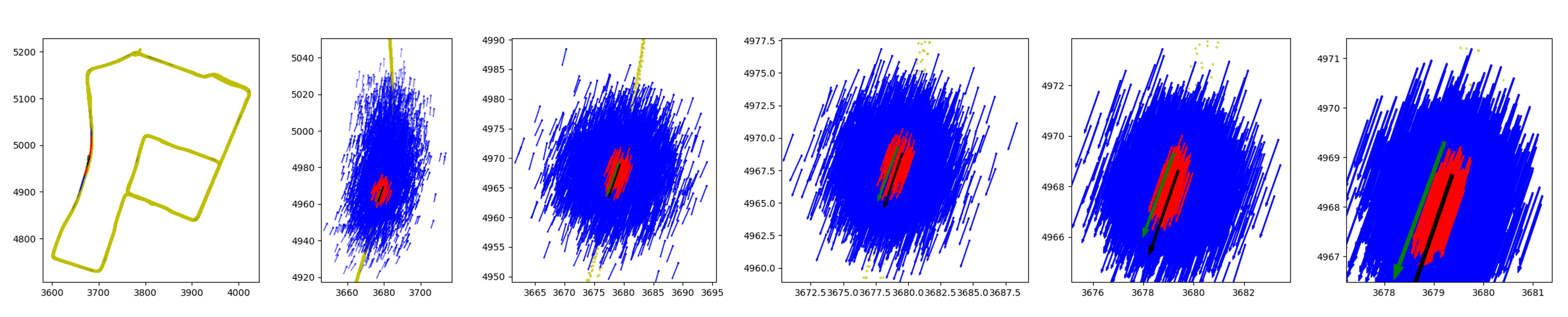}
   \caption{\textbf{Iterative candidates refinement.} At each $k$ step of the localization process, top scored poses are selected to sample the new candidate poses at step $k+1$. From left to right: top scored poses at $k=0$ to $k=5$, yellow points are positions of the training example, blue arrows are pose candidates and red arrows are the selected poses among the candidates.}
   \label{fig:candidates}
\end{figure}

\paragraph{6. Iterative pose refinement:} we repeat $K$ times the evaluation of pose candidates described in steps 3-4-5. After each iteration, the noise vector $v$ is divided by 2, such that new candidates are sampled closer to previous high scores. As a result, we can converge to a precise pose estimate in kilometers-scale maps while only evaluating a limited sparse set of poses. We evaluate each camera frame independently at each time step, however one could use localization priors from previous time steps to reduce the number of iterations in vehicles navigation scenarios. An example of selected poses at each iteration is shown in Fig. \ref{fig:candidates}. By sampling $N$ candidates for initial poses, we preserve a constant memory peak.

\paragraph{7. Pose averaging:} our final camera pose estimate is a weighted average of the 256 pose candidates with higher scores, which exhibits better interpolation properties than selecting the best score pose. We use scores as weighting coefficients and 3D rotation averaging is implemented following~\cite{rotation_averaging}. 

The entire inference procedure requires 1 forward pass on the image encoder and $K$ passes on the pose encoder.

\subsection{Training procedure}

We do not train the system by minimizing the error on the final camera pose estimate. Instead, we apply our loss function directly on the predicted scores. As a result, one training iteration provides supervision on the $K \times N$ image-pose pairs that contains more information than the single localization error. We observed that this property results in superior training efficiency than regression approaches (see \ref{sec:ablation}). We define target scores $s_{t}$ based on translation and rotation distances between the camera pose $p_{I} = (t_{I}, q_{I})$ and the candidate pose $p = (t,q)$:

\begin{equation}
    s_{t}(I,p) = ReLU\left(1 - \lambda_{t} \| t_{i} - t \|_{2} - \lambda_{r}  G(q_{i}, q)\right)
\end{equation}
where $\lambda_{t}$ and $\lambda_{r}$ are weighting parameters set to 5 and 0.1 and $G$ is geodesic distance, defined as the minimal angle between 2 rotations:
\begin{equation}
G(q_1,q_2) = cos^{-1}\left(\frac{(tr(M_{q_1} M_{q_2}^{-1}) - 1)}{2}\right),
\end{equation}
$M_{q}$ being the 3D rotation matrix associated with rotation $q$.

We train $f_{I}$ and $f_{M}$ by computing scores between reference images and pose candidates sampled at $K$ different resolutions as described in section \ref{method_1}. For training purpose, we add to initial poses an uniform noise sampled in $[-v, v]$ as we observed that it reduces overfitting. We also use poses associated with the top target scores in the candidates proposer, in addition with top predicted scores in order to guide training convergence in early iterations.

Finally, our optimization objective is:

\begin{equation}
L = \frac{1}{N} \sum_{k=0}^{K} \sum_{n=0}^{N-1} | s(I,p_{n,k}) - s_{t}(I,p_{n,k}) |
\end{equation}

An analogy can be made with content-based image retrieval~\cite{netvlad,RARS19}: global descriptors are usually trained using image triplets composed of a query image, a positive and a negative example. Positive samples are data close to the query, in metric or semantic domain depending on the final application, and negative samples are images with unrelated content to the query. Global descriptors can be trained by minimizing a triplet margin loss~\cite{netvlad}. In our case, positive examples are the poses with a non-zero score whereas negative examples are candidates farther from the camera pose than an arbitrary threshold. Instead of binary classification (positive or negative example), we rank the relative importance of the positive samples according to their distance to the ground truth label.

%% file: sections/experiments.tex
We compare our approach against recent methods on several datasets covering a wide range of autonomous driving scenarios in large scale outdoor maps. This task is highly challenging due to the dynamic part of outdoor environments (moving objects, illumination, occlusions, etc.). We verify that our formulation enables accurate localization in 9 different large outdoor scenes. Then we show that our method can be naturally extended to multi-map scenarios and we report results using this setup. We also compare the computational efficiency of our method with competitors and finally present an ablation study on hyperparameters of ImPosing. Video displaying trajectories is included in the supplementary material.

\paragraph{Implementation details:} ImPosing is implemented in PyTorch. Images are computed at a small resolution $135\times 240$. The image encoder uses a ResNet34 backbone pretrained on ImageNet. $N=4096$ pose candidates are evaluated at each of the $K=6$ refinement steps. For candidates sampling, the noise vector is set to $v=[8.0m,0.2m,8.0m,1\degree,5\degree,1\degree]$ where y is the altitude axis, and we use 100 GMM components. We train the image encoder and pose encoder for 250 epochs with Adam optimizer at a constant learning rate of $1e^{-4}$. We did not tune these parameters specifically for each scene, suggesting that they should work for any autonomous driving scene. More details are provided in supplementary materials, including datasets configuration.
\paragraph{Baselines:} Our first aim is to compare ImPosing to its direct learning-based methods competitors. We use CoordiNet~\cite{coordinet} that report state-of-the-art results for absolute pose regression on Oxford Dataset as a baseline. We report previously published results on this dataset, and our own implementation for other datasets. We replace the EfficientNet backbone by ResNet34 for a fair comparison with ImPosing. As outlined in section~\ref{sec:related_work}, we share similarities with image retrieval by matching a global descriptor against the map. To compare ImPosing to retrieval, we use NetVLAD~\cite{netvlad} (VGG16 backbone) and Revaud et al.~\cite{RARS19} (GeM pooling, Resnet101 backbone) publicly available implementations\footnote{https://github.com/Nanne/pytorch-NetVlad and https://github.com/naver/deep-image-retrieval}. Full sized images are used to compute global image descriptors followed by cosine similarity for features comparison, then we perform pose averaging on poses of top 20 database images as in~\cite{Sattler2019}. Scene coordinate regression~\cite{brachmann2019esac,dsac*} can not scale to large environments thus is not considered for evaluation. We did not conduct experiments with structure-based methods~\cite{sarlin2019coarse,sarlin2021pixloc,DA4AD_2020_ECCV}. These methods are more accurate than ours thanks to geometric reasoning with a 3D model, but also operate at a different computation scale than ours (see figure~\ref{fig:memory}) making embedded deployment difficult. In scenarios where it can be afforded, ImPosing can be considered as a coarse localization step, followed by refinement with a 3D model, similar to HLoc\cite{sarlin2019coarse} architecture.

\subsection{Single scene localization}

\input{tables/oxford_table}

\paragraph{Oxford RobotCar~\cite{RobotCarDatasetIJRR}} contains images recorded by a vehicle in Oxford over a year. We reproduce experiments commonly reported for learning-based methods~\cite{coordinet,AtLoc,graph_neural_network}: we evaluate on the \textit{Loop} and \textit{Full} scenes, using only 2 sequences for training. Results are reported in Table~\ref{tab:oxford-lake}.

First we observe that image retrieval performs better than pose regression. Previous learning-based methods struggle due to the low-data regime~\cite{coordinet,lens} and the decrease of the regression accuracy in large maps. Oxford city is an environment with rich features similar to visual place recognition training datasets, that make NetVLAD~\cite{netvlad} and GeM~\cite{RARS19} strong baselines in this scenario. ImPosing exhibits state-of-the art accuracy on Oxford Loop scene, as well as the best mean error in average. These results are obtained by reducing a lot the number of large failure cases that occur with prior methods.

We also observe that despite newly provided RTK ground truth provided by the authors~\cite{RCDRTKArXiv}, the reference poses are largely inaccurate in some areas. As a result, evaluation metrics are not significant at a centimeter level and models training might be impacted by this erroneous pose labels. For this reason, we conduct a benchmark on two recently released datasets with more reliable ground-truth.

\paragraph{Daoxiang Lake~\cite{DA4AD_2020_ECCV}} has been collected in a 12km loop in Beijing during 4 months. 8 recordings are available, we use 7 for training and 1 for testing with images from the front camera only. This scene contains the largest map and training dataset of our experiments. Median and mean errors are shown in Table~\ref{tab:oxford-lake}. Daoxiang Lake is a more challenging dataset than Oxford because of repetitive areas with few discriminative features and various environments (urban, peri-urban, highways, nature, etc.). As a result, image retrieval performs worse than pose regression. ImPosing is way more accurate and exhibits a median error 4 times smaller than competitors.

\paragraph{4 seasons~\cite{wenzel2020fourseasons}} contains data recorded in Munich area in various scenes (city, residential neighborhoods, countrysides) with varying seasonal conditions. We selected 6 scenes where at least 3 different recordings are provided: we use 1 for testing and others as training images. This benchmark is highly challenging due to extreme appearance changes between sequences, small data regime for some scenes, featureless environments (see illustration in supplementary materials) and kilometers-scale maps. Results are reported in table~\ref{tab:4seasons}.

\input{tables/4seasons_table}

First, absolute localization accuracy is very heterogeneous between different scenes. We note that scenes with few training images are the most challenging. In particular, \textit{Countryside} include navigation around fields and \textit{City Loop} is a 10km map where the training dataset is composed of a winter sequence with snow and a rainy sequence with blur on camera lens. In these extreme cases, both pose regression and image retrieval fail to estimate reliable poses, whereas ImPosing is able to provide a coarse localization. With sufficiently large training datasets, our method still exhibits the more precise pose estimation.




\subsection{Multi-scene localization}

Learning-based methods for relocalization require scene specific training, inducing heavy computation for potential deployment in several areas at a large scale. Recent work~\cite{blanton2020extending,shavit2021learning} has extended absolute pose regression to multi-scene scenarios. The core idea is to train a system with images from several maps while sharing image encoder parameters that could learn to extract features in a generic way. As our method separate image and map representation, ImPosing naturally extends to multi scenes scenarios. To adapt ImPosing to a multi-map scenario, we perform the following modifications: the image encoder backbone is shared between all maps, whereas one specific pose encoder is learned for each scene. We also learn scene specific parameters for the final linear layer of the image encoder, to facilitate image features projection to the desired map representation.
We train a multi-scene model on the 6 maps of 4 seasons~\cite{wenzel2020fourseasons}. Results are reported in Table~\ref{tab:4seasons}. The model has been trained for 20 epochs only because of computational constraints, but still outperform all competitors except single scenes ImPosing models. While the convergence for a single scene is slower in the multimap formulation (but training a multiscene on $n$ maps is faster than performing $n$ different trainings on each map, see supplementary materials), it enables to localize in huge areas with minimal memory storage requirements (see section~\ref{efficiency}).



\subsection{Efficiency comparison}
\label{efficiency}

\input{tables/comparison_table}

\begin{figure}[h]
\centering
   \includegraphics[width=\linewidth]{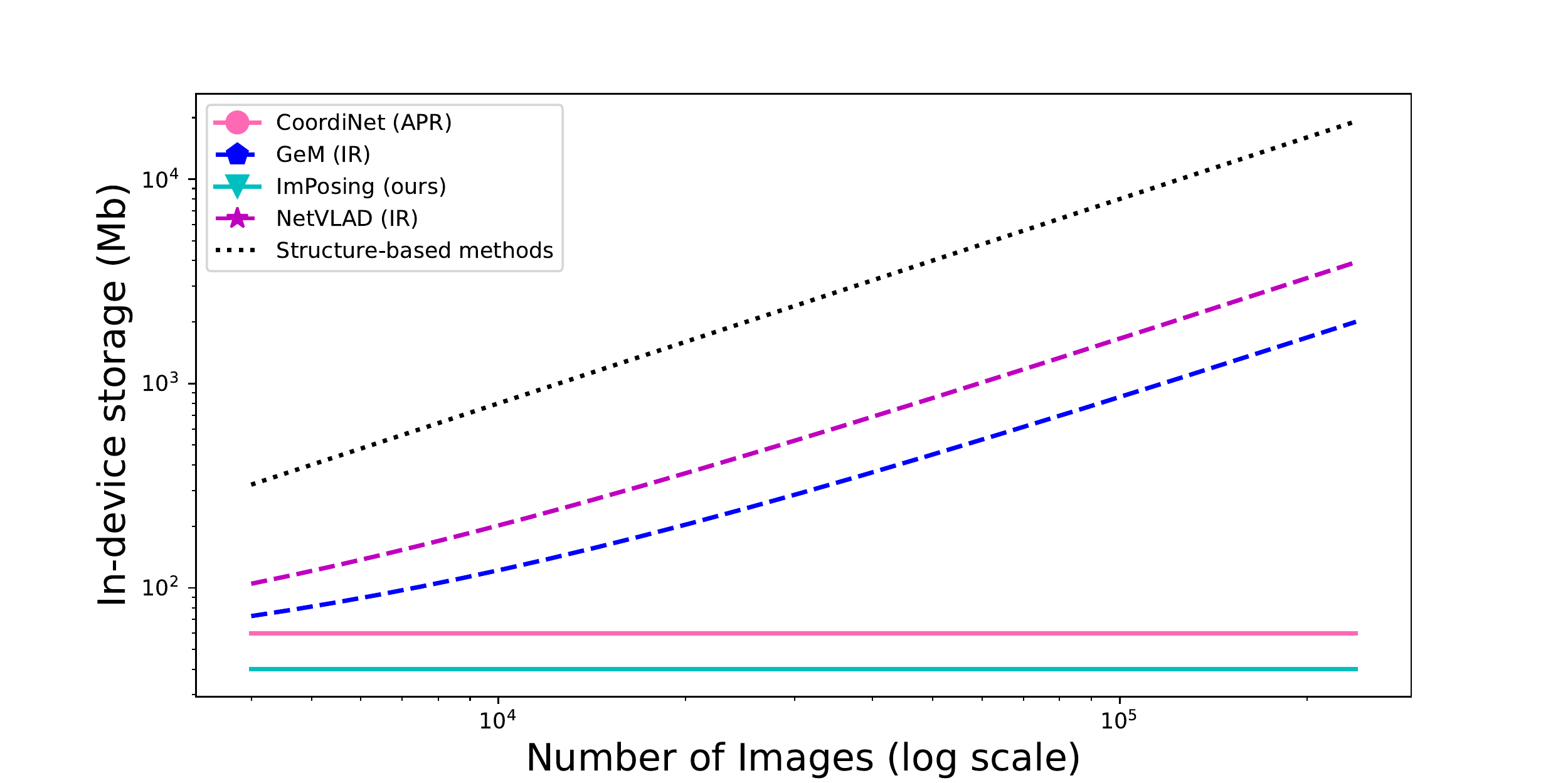}
   \caption{\label{fig:memory} \textbf{In-device memory usage.} Structure-based methods (black) and image retrieval (blue and purple) use more memory when the reference dataset grows whereas pose regression methods and ImPosing (pink and cyan) storage requirement does not depend on dataset size.}
\end{figure}

\paragraph{Storage footprint.} Our method only needs to store neural networks weights and initial pose candidates in device. It represents 23MB for the image encoder, less than 1MB for the pose encoder and 1MB for the initial poses candidates. We also report in figure~\ref{fig:memory} the scaling law of memory footprint w.r.t. reference database size for different classes of visual localization methods. This is an important aspect in autonomous driving scenarios where large amounts of data are available. For a given map, learning-based methods have a constant memory requirement because the map information is embedded in the networks weights. To estimate storage requirement of retrieval methods, we consider the size of the database image descriptor (2048 for GeM and 4096 for NetVLAD) along with the size of the image encoder. Storage requirement of retrieval methods exceed 1 GB for large scale scene with more than 100k reference images. To estimate the memory requirement of structure-based methods we consider the numbers given in~\cite{sarlin2019coarse}: a 3D model built from 4328 images is composed of 685k 3D points. If we consider one local descriptor of size 128 by 3D points, we can derive a linear rule to determine the 3D model size according to the number of reference images. This is a rough estimation but we can estimate that structure based method require at least 3 times more storage capacity than image retrieval methods. Compressing techniques exist to make these methods more tractable~\cite{hyperpoints,hybrid_compression}, however compressed maps still represent gigabytes and are less accurate.


\paragraph{Computational complexity.} Our algorithm complexity depends on the image encoder backbone (3.6 billion FLOPs for ResNet34) and the hierarchical decoding process with the pose encoder. With the default hyperparameters, it involves 4.8 billion FLOPs. We measured a total inference time of 41ms for a single image using a NVIDIA RTX 2080 GPU. The complexity is linear w.r.t. the number of refinements $K$, the number of pose candidates $N$ and the number of layers in the MLP. It is quadratic w.r.t. the latent dimension $D$. It should be noted that parallel computations reduce the impact of $N$ and $D$ on the inference time. Considering these properties and the ablations provided in \ref{sec:ablation}, one can choose the corresponding hyperparameters that match its computational requirements.

\input{tables/ablation}
\paragraph{Summary.} ImPosing exhibits very compact storage requirements and fast inference time coupled with state-of-the-art accuracy. Notably, neither memory footprint and computational complexity depends on the number of images in the reference database, which is a great advantage over image retrieval methods~\cite{netvlad,RARS19}. We also observe empirically that our method converges approximately 2 times faster than pose regression competitors~\cite{coordinet} w.r.t. the number of training iterations (see figure~\ref{fig:ablation}).

\subsection{Ablation study}
\label{sec:ablation}

We report the influence of several hyperparameters on the localization accuracy of ImPosing in figure~\ref{fig:ablation}. We evaluate the number of refinement steps $K$, the number of pose candidates $N$ and the number of best candidates used for pose averaging. We use the model trained on Daoxiang Lake and change the parameters at test time. Increasing the number of refinements and candidates improves localization accuracy, at the cost of a higher computational cost. We use a reasonable trade-off with $K=6$ and $N=4096$ as our default setup. We observe that pose averaging has a positive impact on accuracy, but the number of selected candidates is not critical. Additional ablation studies on number of layers in the pose encoder and computation of the similarity score are provided in supplementary materials.

%% file: tables/oxford_table.tex
\begin{table*}[h]
\centering
\scriptsize


\renewcommand{\arraystretch}{1.2}
\begin{tabular}{@{} l l ll c ll c l @{}}
\toprule
&& \multicolumn{2}{c}{Pose regression} && \multicolumn{2}{c}{Image retrieval} && \multirow{2}{*}{ImPosing} \\
\cmidrule{3-4} \cmidrule{6-7}
Dataset && CoordiNet~\cite{coordinet} & AtLoc~\cite{AtLoc} && NetVLAD~\cite{netvlad} & GeM~\cite{RARS19} &&  \\
\hline
\multirow{2}{*}{Oxford Full}
& Median & 3.55m/1.1° & 11.1m/5.3° && 1.42m/1.4° & \textbf{1.36m/1.3°} && 1.90m/1.3° \\
& Mean   & 14.96m/5.7° & 29.6m/12.4° && 4.47m/2.4° & \textbf{3.49m/2.3°} && 4.25m/4.3° \\ \hdashline
\multirow{2}{*}{Oxford Loop}
& Median & 2.27m/0.9° & 5.36m/2.1° && 2.16m/1.1° & 2.39m/1.0° && \textbf{1.93m/1.0°} \\
& Mean & 4.15m/1.4° &  8.73m/4.6° && 4.16m/1.9° & 6.92m/3.1° &&  \textbf{3.03m/1.8°} \\ \hdashline
\multirow{2}{*}{Average}       
& Median & 2.91m/1.0° & 8.23m/3.7° && \textbf{1.79m}/1.2° & 1.88m/\textbf{1.1°} && 1.92m/\textbf{1.1°} \\
& Mean & 9.56m/3.4° & 19.17m/8.5° && 4.32m/\textbf{2.1°} & 5.20m/2.7° && \textbf{3.64m}/3.0° \\
\hline
\multirow{2}{*}{Daoxiang Lake}
& Median & 6.82m/0.4° & -- && 8.92m/0.8° & 27.13m/1.1° && \textbf{1.62m/0.3°} \\
& Mean & 25.18m/1.0° & -- && 152.2m/15.5° & 328.8m/19.5° && \textbf{8.40m/0.5°} \\
\bottomrule
\end{tabular}
\caption{\label{tab:oxford-lake} Localization error on Oxford RobotCar and Daoxiang Lake datasets.}
\end{table*}

%% file: tables/4seasons_table.tex
\begin{table*}[h]
\centering
\tiny

\renewcommand{\arraystretch}{1.2}
\begin{tabular}{@{} l c lll c ll c l c ll @{}}
\toprule
&& \multicolumn{3}{c}{Dataset details} && \multicolumn{2}{c}{Image retrieval} && \multirow{2}{*}{CoordiNet~\cite{coordinet}} && \multicolumn{2}{c}{ImPosing} \\
\cmidrule{3-5} \cmidrule{7-8} \cmidrule{12-13}
 && Road length & Runs & Images && NetVLAD \cite{netvlad} & GeM~\cite{RARS19} && && Single sc. & Multi sc. \\
 \hline
Neighborhood       && 2000 & 6 & 16520 && 0.72m/0.9° & 0.69m/0.9° && 0.74m/\textbf{0.6°} && \textbf{0.53m}/0.7° & 0.82m/1.0° \\
Office loop        && 2600 & 5 & 20915 && 6.85m/3.0° & 6.39m/2.8° && 6.25m/1.5° && \textbf{0.99m/1.1°} & 1.58m/1.3° \\
Countryside        && 6200 & 3 & 19804 && 32.24m/1.2° & 30.87m/1.3° && 47.33m/2.9° && \textbf{2.61m/0.9°} & 5.46m/1.1° \\
Bus. campus        && 1000 & 2 & 6132  && 1.19m/1.3° & 1.96m/\textbf{1.2°} && 22.57m/6.0° && \textbf{1.16m}/1.3° & 1.70m/1.6°\\
City loop          && 10000 & 2 & 17427&& 61.60m/3.5° & 317.4m/6.9° && 584.4m/14.4° && \textbf{5.32m/2.4°} & 10.53m/2.5°\\
Old Town           && 4500 & 3 & 13959 && 3.45m/\textbf{1.2°} & 4.46m/1.6° && 50.83m/3.8° && \textbf{2.59m/1.2°} & 3.71m/1.3°\\
Average            && - & - & -        && 17.67m/1,8° & 60.30m/2,4° && 118.7m/4.9° && \textbf{2.2m/1.3°} & 3.97m/1.5° \\ 
\bottomrule
\end{tabular}
\caption{\label{tab:4seasons} Median localization error on 4Seasons dataset.}
\end{table*}

%% file: tables/comparison_table.tex
\begin{table}[b]
\resizebox{\linewidth}{!}{
\begin{tabular}{|l|l|l|l|l|}
\hline
    \textbf{Algorithms} & \textbf{In device storage} & \textbf{Scalability} & \textbf{Latency} & \textbf{Accuracy} \\ \hline
    IR+2D-3D matching & 3D model + IR DB + NN (5-100GBs) & High & Low & High  \\ \hline
    IR+Relative PR & IR DB with images + NN (5-100GBs) & High & Low & Medium  \\ \hline
    IR & IR DB + NN (2-50GBs)& High & Medium & Low  \\ \hline
    APR & NN ($\approx$ 25MB) & Medium & High & Low  \\ \hline
    SCR & NN ($\approx$ 25MB) & Low & High & High  \\ \hline
    \textbf{ImPosing (ours)} & NN (25MB) & High & High & Medium  \\ \hline

\end{tabular}
}
\caption{\label{tab} \textbf{Qualitative comparison between methods.} We compare the properties of visual localization class of methods w.r.t. storage requirement, capability to operate in large maps (scalability), latency and accuracy. \textit{IR} stands for Image Retrieval, \textit{PR} for Pose Regression, \textit{SCR} for Scene Coordinate Regression, \textit{DB} for database and \textit{NN} for neural networks weights. Storage of IR databases are detailed in~\cite{song2022dalg}.}
\end{table}

%% file: tables/ablation.tex
\begin{table*}[h!]
\centering
\begin{minipage}[t]{0.2\textwidth}
\begin{tikzpicture}[xscale=0.5,yscale=0.5]
\begin{axis}[
    xlabel={\Large Number of refinements K},
    ylabel={\Large Test median error},
    xmin=0, xmax=10,
    ymin=0, ymax=4,
    xtick={0,1,2,3,4,5,6,7,8,9},
    ytick={0,1.0,2.0,3.0,4.0},
    legend pos=north east,
    ymajorgrids=true,
    grid style=dashed,
]

\addplot[
    color=red,
    mark=*,
    ]
    coordinates {
    (0,46)(1,25.15)(2,3.34)(3,1.97)(4,1.75)(5,1.66)(6,1.62)(7,1.60)(8,1.59)(9,1.59)
    };

\addplot[
    color=cyan,
    mark=*,
    ]
    coordinates {
    (0,1.78)(1,0.86)(2,0.47)(3,0.34)(4,0.30)(5,0.29)(6,0.29)(7,0.29)(8,0.29)(9,0.30)
    };
    
\legend{Translation (m),Rotation (°)}
\end{axis}
\end{tikzpicture}
\end{minipage}
\hspace{0.05\textwidth}
\begin{minipage}[t]{0.2\textwidth}
\begin{tikzpicture}[xscale=0.5,yscale=0.5]
\begin{axis}[
    xlabel={\Large Number of candidates N},
    xmin=0, xmax=8192,
    ymin=0, ymax=4,
    xtick={1024, 2048, 4096, 8192},
    ytick={0,1.0,2.0,3.0,4.0},
    legend pos=north east,
    ymajorgrids=true,
    grid style=dashed,
]

\addplot[
    color=red,
    mark=*,
    ]
    coordinates {
    (1024,3.15)(2048,2.02)(4096,1.62)(8192,1.54)
    };

\addplot[
    color=cyan,
    mark=*,
    ]
    coordinates {
    (1024,0.43)(2048,0.33)(4096,0.29)(8192,0.28)
    };
    
\legend{Translation (m),Rotation (°)}
\end{axis}
\end{tikzpicture}
\end{minipage}
\hspace{0.05\textwidth}
\begin{minipage}[t]{0.2\textwidth}
\begin{tikzpicture}[xscale=0.5,yscale=0.5]
\begin{axis}[
    xlabel={\Large Number of pose averaging candidates},
    xmin=0, xmax=512,
    ymin=0, ymax=4,
    xtick={1, 64,128,256,512},
    ytick={0,1.0,2.0, 3.0, 4.0},
    legend pos=north east,
    ymajorgrids=true,
    grid style=dashed,
]

\addplot[
    color=red,
    mark=*,
    ]
    coordinates {
    (1,1.99)(64,1.63)(128,1.63)(256,1.62)(512,1.61)
    };

\addplot[
    color=cyan,
    mark=*,
    ]
    coordinates {
    (1,0.38)(64,0.29)(128,0.29)(256,0.29)(512,0.29)
    };
    
\legend{Translation (m),Rotation (°)}
\end{axis}
\end{tikzpicture}
\end{minipage}
\hspace{0.05\textwidth}
\begin{minipage}[t]{.2\textwidth}
\input{tables/training_efficiency}    

\end{minipage}
\hspace{0.05\textwidth}
\captionof{figure}{\label{fig:ablation} \textbf{From left to right:} median localization errors depending on number of refinements, pose candidates, and final number averaged poses. Training time comparison between pose regression~\cite{coordinet} and ImPosing.}
\end{table*}

%% file: tables/training_efficiency.tex
\begin{tikzpicture}[xscale=0.5,yscale=0.5]
\begin{axis}[
    xlabel={\Large Training iterations ($\times 10^3$)},
    xmin=0, xmax=20,
    ymin=0, ymax=10,
    xtick={0,5,10,15,20},
    ytick={0,1.0,2.0,3.0,4.0,5,6,7,8,9},
    legend pos=north east,
    ymajorgrids=true,
    grid style=dashed,
]

\addplot[
    color=red,
    ]
    coordinates {
(0.412,	6.240188599)
(0.825,	2.670292854)
(1.238,	1.469020247)
(1.651,	1.191830397)
(2.064,	1.014350057)
(2.477,	1.114924908)
(2.890,	0.88902384)
(3.303,	0.929884791)
(3.716,	0.880360305)
(4.129,	0.914007843)
(4.542,	0.933842242)
(4.955,	0.878960192)
(5.368,	0.769154787)
(5.781,	0.832954586)
(6.194,	0.810410559)
(6.607,	0.781660199)
(7.020,	0.762929916)
(7.433,	0.797301888)
(7.846,	0.76713413)
(8.259,	0.739594817)
(8.672,	0.814575434)
(9.085,	0.768811405)
(9.498,	0.747829854)
(9.911,	0.711111784)
(10.324,	0.747537374)
(10.737,	0.76447165)
(11.150,	0.732137322)
(11.563,	0.767292142)
(11.976,	0.698736072)
(12.389,	0.734788239)
(12.802,	0.717812121)
(13.215,	0.75308764)
(13.628,	0.683090568)
(14.041,	0.677256763)
(14.454,	0.678701043)
(14.867,	0.692996442)
(15.280,	0.644741595)
(15.693,	0.652901709)
(16.106,	0.682376742)
(16.519,	0.648912847)
(16.932,	0.638795555)
(17.345,	0.659274101)
(17.758,	0.650959551)
(18.171,	0.655239344)
(18.584,	0.650561094)
(18.997,	0.662347555)
(19.410,	0.654489875)
(19.823,	0.677235246)
(20.236,	0.64382416)
(20.649,	0.668699503)
    };

\addplot[
    color=orange,
    ]
    coordinates {
(0.492,	12.44909477)
(0.574,	7.756752968)
(0.984,	5.748052597)
(1.230,	5.4037838)
(1.312,	4.395642281)
(1.558,	2.441093922)
(1.640,	2.322518349)
(1.722,	2.145583153)
(2.050,	2.258273602)
(3.034,	1.704257965)
(3.362,	1.718227983)
(3.526,	1.684372663)
(3.690,	1.518226147)
(3.772,	1.509226322)
(3.936,	1.599899054)
(4.428,	1.648850679)
(4.510,	1.607573509)
(4.592,	1.477819443)
(4.674,	1.578014016)
(4.756,	1.596609116)
(5.248,	1.418224335)
(5.412,	1.417478919)
(5.576,	1.490007639)
(5.658,	1.542612553)
(5.904,	1.418759346)
(5.986,	1.439119816)
(6.068,	1.433094025)
(6.150,	1.331451654)
(6.560,	1.337465286)
(6.970,	1.325685501)
(7.216,	1.463678241)
(7.298,	1.342944145)
(7.790,	1.307371616)
(8.446,	1.479376674)
(8.692,	1.243725896)
(8.774,	1.356505275)
(9.348,	1.20983243)
(9.594,	1.293931484)
(9.758,	1.238755465)
(10.332,	1.116316557)
(10.414,	1.469472766)
(10.660,	1.26887393)
(10.742,	1.340078592)
(11.316,	1.216734409)
(11.480,	1.189817667)
(11.644,	1.18289423)
(11.726,	1.242947578)
(11.808,	1.217341661)
(12.382,	1.199080348)
(12.792,	1.195028424)
(12.956,	1.164628386)
(13.038,	1.070818543)
(13.366,	1.198552608)
(13.776,	1.232396722)
(13.858,	1.142018676)
(14.104,	1.197324395)
(14.596,	1.212868094)
(14.678,	1.138986826)
(14.842,	1.157865047)
(15.088,	1.307216167)
(15.252,	1.254741669)
(15.334,	1.306593418)
(15.498,	1.169724464)
(15.662,	1.109075069)
(15.744,	1.214944124)
(16.318,	1.386652946)
(16.974,	1.115156889)
(17.056,	1.327528715)
(17.138,	1.208262444)
(17.466,	1.078688622)
(18.286,	1.089944363)
(18.368,	1.161625862)
(18.532,	1.140420318)
(18.860,	1.149354339)
(19.106,	1.245787859)
(19.188,	1.206497669)
(19.270,	1.268007159)
(19.516,	1.106123924)
(20.500,	1.156291723)
    };
    
\legend{ImPosing (ours), CoordiNet~\cite{coordinet}}
\end{axis}
\end{tikzpicture}

%% file: sections/discussion.tex
\paragraph{What does the pose encoder learn?}
In the pose regression approach, image and camera pose are connected by being the respective input and output of a single feedforward neural network. This formulation entangles features extraction, map memorization and camera pose prediction in a single model. While deep neural networks are known to perform well for the first, they have been observed to be inaccurate for pose prediction~\cite{Sattler2019}. Our solution circumvents this problem by "inverting" the decoder layers with the pose encoder. We don't try to predict the pose from features but to connect a given pose to its respective latent features. We let the network learn the optimal latent space to connect images and camera poses, with a single constraint: pose candidates close to the actual camera pose must have a vector relatively similar to the image representation. This property enables to search the best pose candidates in a coarse to fine manner, and interpret the resulting scores has a multimodal distribution of positions across the map. We provide visualizations of these distributions and of the latent space structure in supplementary materials.

\paragraph{Benefits, limitations and future work.} Our method keeps the main advantages of direct learning-based methods: we obtain the pose efficiently with neural networks inference, we do not use a 3D model of the scene or a retrieval database, resulting in a very compact memory footprint.
We observe that the accuracy our method highly depends on the quantity of training data available. Similar to regression, our method does not extrapolate to camera positions far from trainings examples. However, recent approaches has shown that these limitations can be overcome with synthetic datasets~\cite{lens}. Moreover, in the driving scenario, a coarse localization estimate can be sufficient because horizontal localization (road lane) can be recovered thanks to perception~\cite{roadmap}.
The new paradigm we propose could be improved in many ways. It includes exploring better architectures for the pose encoder, inspired from recent work on coordinate-based representations~\cite{zhu_cvpr}. Another interesting direction is to extend the implicit map representation to local features instead of global image signatures, by finding a way to represent implicitly a 3D model.

%% file: sections/conclusion.tex
We have proposed a new formulation for visual localization that perform state-of-the art accuracy for direct learning-based methods in large environments. By using an implicit representation of the map, we connect camera poses and image features in a latent high dimensional manifold well suited for localization. We have shown that with a simple pose candidates sampling procedure, we are able to estimate the absolute pose of an image. Our proposal can be directly applied in autonomous driving systems, by providing an efficient and accurate image-based localization algorithm that can operate at large scales in real-time. We believe that, beyond our work, implicit scene representations, by their ability to model complex continuous signals in a fixed size neural network, are a promising research direction for camera pose estimation. 

%% file: tables/pose_encoder_ablation.tex
\begin{table}[h!]
\centering
\begin{minipage}[t]{0.3\textwidth}
\begin{tikzpicture}[xscale=0.7,yscale=0.7]
\begin{axis}[
    xlabel={\Large Neighborhood: pose encoder layers},
    ylabel={\Large Test median error},
    xmin=1, xmax=10,
    ymin=0, ymax=2,
    xtick={2,4,8},
    ytick={0,0.5,1.0,1.5,2.0},
    legend pos=north east,
    ymajorgrids=true,
    grid style=dashed,
]

\addplot[
    color=red,
    mark=*,
    ]
    coordinates {
    (2,.48)(4,.53)(8,.54)
    };

\addplot[
    color=cyan,
    mark=*,
    ]
    coordinates {
    (2,.9)(4,.7)(8,1.0)
    };
    
\legend{Translation (m),Rotation (°)}
\end{axis}
\end{tikzpicture}
\end{minipage}
\hspace{0.2\textwidth}
\begin{minipage}[t]{0.3\textwidth}
\begin{tikzpicture}[xscale=0.7,yscale=0.7]
\begin{axis}[
    xlabel={\Large Countryside: pose encoder layers},
    xmin=1, xmax=10,
    ymin=0, ymax=4,
    xtick={2,4,8},
    ytick={0,0.5,1.0,1.5,2.0,2.5,3,3.5,4},
    legend pos=north east,
    ymajorgrids=true,
    grid style=dashed,
]

\addplot[
    color=red,
    mark=*,
    ]
    coordinates {
    (2,2.15)(4,2.61)(8,2.95)
    };

\addplot[
    color=cyan,
    mark=*,
    ]
    coordinates {
    (2,.9)(4,.9)(8,1.0)
    };
    
\legend{Translation (m),Rotation (°)}
\end{axis}
\end{tikzpicture}
\end{minipage}
\hspace{0.05\textwidth}
\captionof{figure}{\label{fig:ablation} \textbf{Localization accuracy depending on pose encoder capacity}}
\end{table}

%% file: tables/matching_module_ablation.tex
\begin{figure}
    \centering
    \begin{tikzpicture}
    \begin{axis}[major x tick style = transparent,
        ybar,
        ylabel={\Large Test median error},
        ymin=0, ymax=2,
        enlarge x limits=0.25,
        bar width=20pt,
        ymajorgrids = true,
        symbolic x coords={CS,L2,MLP},
        ytick={0,0.5,1.0,1.5,2.0},
        legend pos=north east,
        grid style=dashed,
        xtick = data
    ]

    \addplot[color=red, fill=red]
        coordinates {(CS,.53)(L2,.81)(MLP,.72)};
    
    \addplot[color=cyan, fill=cyan]
        coordinates {
        (CS,.7)(L2,1.4)(MLP,1.2)
        };
        
    \legend{Translation (m),Rotation (°)}
    \end{axis}
    \end{tikzpicture}
    \caption{\label{fig:match_mod} \textbf{Localization accuracy depending on similarity score computation.} CS stands for cosine similarity, L2 for the distance-based similarity and MLP for the learned score computation}
    \label{fig:my_label}
\end{figure}

%% file: tables/oxford_split.tex
\begin{center}
\begin{tabular}{|c|c|c|}
\hline
 & Oxford Loop & Oxford Full \\
\hline
 Training set & 2014-06-26-09-24-58 & 2014-11-28-12-07-13 \\
 & 2014-06-23-15-41-25 & 2014-12-02-15-30-08 \\
 \hline
 Test set & 2014-06-26-08-53-56 & 2014-12-09-13-21-02\\
 & 2014-06-23-15-36-04 & \\
\hline
\end{tabular}
\end{center}

%% file: tables/daoxianglake_split.tex
\begin{center}
    \begin{tabular}{|c|c|}
    \hline
 & Daoxiang Lake dataset \\
 \hline
 & 20191216123346 \\
 & 20191130112819 \\
 & 20191025104732 \\
 Training set & 20191021162130 \\
 & 20191014142530 \\
 & 20190924124848 \\
 & 20190918143332 \\
 \hline
 Test set & 20191225153609 \\
 \hline
\end{tabular}
\end{center}

%% file: tables/4seasons_split.tex
\scalebox{0.4}{
\begin{tabular}{|c|c|c|c|c|c|c|}
\hline
 & Neighborhood & Office Loop & Countryside & Bus. campus & City Loop & Old Town \\
 \hline
 Train & 2020-03-26\_13-32-55 & 2020-03-24\_17-36-22 & 2020-04-07\_11-33-45 & 2020-10-08\_09-30-57 & 2020-12-22\_11-33-15 & 2020-10-08\_11-53-41\\
 & 2020-10-07\_14-47-515 & 2020-03-24\_17-45-31 & 2020-06-12\_11-26-43 & 2021-01-07\_13-12-23 & 2021-01-07\_14-36-17 & 2021-01-07\_10-49-45 \\
 & 2020-10-07\_14-53-52 & 2020-04-07\_10-20-32 & 2021-01-07\_13-30-07 & & & 2021-05-10\_21-32-00 \\
 & 2020-12-22\_11-54-24 & 2020-06-12\_10-10-57 & & & & \\
 & 2021-02-25\_13-25-15 & 2021-01-07\_12-04-03 & & & & \\
 & 2021-05-10\_18-02-12 & & & & & \\
 \hline
 Test & 2021-05-10\_18-32-32 & 2021-02-25\_13-51-57 & 2020-10-08\_09-57-28 & 2021-02-25\_14-16-43 & 2021-02-25\_11-09-49 & 2021-02-25\_12-34-08 \\
 \hline
\end{tabular}
}

%% file: egpaper.bbl
\begin{thebibliography}{10}\itemsep=-1pt

\bibitem{netvlad}
R. Arandjelovi\'c, P. Gronat, A. Torii, T. Pajdla, and J. Sivic.
\newblock {NetVLAD}: {CNN} architecture for weakly supervised place
  recognition.
\newblock In {\em IEEE Conference on Computer Vision and Pattern Recognition},
  2016.

\bibitem{Atzmon_2020_CVPR}
Matan Atzmon and Yaron Lipman.
\newblock Sal: Sign agnostic learning of shapes from raw data.
\newblock In {\em IEEE/CVF Conference on Computer Vision and Pattern
  Recognition (CVPR)}, June 2020.

\bibitem{relocnet}
Vassileios Balntas, Shuda Li, and Victor Prisacariu.
\newblock Relocnet: Continuous metric learning relocalisation using neural
  nets.
\newblock In {\em The European Conference on Computer Vision (ECCV)}, September
  2018.

\bibitem{blanton2020extending}
Hunter Blanton, Connor Greenwell, Scott Workman, and Nathan Jacobs.
\newblock Extending absolute pose regression to multiple scenes.
\newblock In {\em Proceedings of the IEEE/CVF Conference on Computer Vision and
  Pattern Recognition Workshops}, pages 38--39, 2020.

\bibitem{brachmann2017dsac}
Eric Brachmann, Alexander Krull, Sebastian Nowozin, Jamie Shotton, Frank
  Michel, Stefan Gumhold, and Carsten Rother.
\newblock Dsac-differentiable ransac for camera localization.
\newblock In {\em Proceedings of the IEEE conference on computer vision and
  pattern recognition}, pages 6684--6692, 2017.

\bibitem{Brachmann_2018_CVPR}
Eric Brachmann and Carsten Rother.
\newblock Learning less is more - 6d camera localization via 3d surface
  regression.
\newblock In {\em Proceedings of the IEEE Conference on Computer Vision and
  Pattern Recognition (CVPR)}, 2018.

\bibitem{brachmann2019esac}
Eric Brachmann and Carsten Rother.
\newblock Expert sample consensus applied to camera re-localization.
\newblock In {\em ICCV}, 2019.

\bibitem{dsac*}
Eric Brachmann and Carsten Rother.
\newblock Visual camera re-localization from {RGB} and {RGB-D} images using
  dsac.
\newblock In {\em IEEE Transactions on Pattern Analysis and Machine
  Intelligence}, pages 1--1, 2021.

\bibitem{hybrid_compression}
Federico Camposeco, Andrea Cohen, Marc Pollefeys, and Torsten Sattler.
\newblock Hybrid scene compression for visual localization.
\newblock In {\em 2019 IEEE/CVF Conference on Computer Vision and Pattern
  Recognition (CVPR)}, pages 7645--7654, 2019.

\bibitem{vidloc}
Ronald Clark, Sen Wang, Andrew Markham, Niki Trigoni, and Hongkai Wen.
\newblock Vidloc: A deep spatio-temporal model for 6-dof video-clip
  relocalization.
\newblock In {\em Proceedings of the IEEE Conference on Computer Vision and
  Pattern Recognition}, pages 6856--6864, 2017.

\bibitem{detone18superpoint}
Daniel DeTone, Tomasz Malisiewicz, and Andrew Rabinovich.
\newblock Superpoint: Self-supervised interest point detection and description.
\newblock In {\em CVPR Deep Learning for Visual SLAM Workshop}, 2018.

\bibitem{camnet}
Mingyu Ding, Zhe Wang, Jiankai Sun, Jianping Shi, and Ping Luo.
\newblock Camnet: Coarse-to-fine retrieval for camera re-localization.
\newblock In {\em 2019 IEEE/CVF International Conference on Computer Vision
  (ICCV)}, pages 2871--2880, 2019.

\bibitem{ding2019camnet}
Mingyu Ding, Zhe Wang, Jiankai Sun, Jianping Shi, and Ping Luo.
\newblock Camnet: Coarse-to-fine retrieval for camera re-localization.
\newblock In {\em Proceedings of the IEEE/CVF International Conference on
  Computer Vision}, pages 2871--2880, 2019.

\bibitem{RANSAC}
M. Fischler and R. Bolles.
\newblock Random sample consensus: A paradigm for model fitting with
  applications to image analysis and automated cartography.
\newblock {\em Communications of the ACM}, 24(6):381--395, 1981.

\bibitem{GARL17}
A. Gordo, J. Almazan, J. Revaud, and D. Larlus.
\newblock End-to-end learning of deep visual representations for image
  retrieval.
\newblock {\em IJCV}, 2017.

\bibitem{Local_Implicit_Grid_CVPR20}
Chiyu~Max Jiang, Avneesh Sud, Ameesh Makadia, Jingwei Huang, Matthias Nießner,
  and Thomas Funkhouser.
\newblock Local implicit grid representations for 3d scenes.
\newblock In {\em Proceedings IEEE Conf. on Computer Vision and Pattern
  Recognition (CVPR)}, 2020.

\bibitem{bayesianposenet}
Alex Kendall and Roberto Cipolla.
\newblock Modelling uncertainty in deep learning for camera relocalization.
\newblock In {\em 2016 IEEE international conference on Robotics and Automation
  (ICRA)}, page 4762–4769. IEEE Press, 2016.

\bibitem{PoseNet}
A. {Kendall}, M. {Grimes}, and R. {Cipolla}.
\newblock Posenet: A convolutional network for real-time 6-dof camera
  relocalization.
\newblock In {\em 2015 IEEE International Conference on Computer Vision
  (ICCV)}, pages 2938--2946, 2015.

\bibitem{rpr_laskar}
Zakaria Laskar, Iaroslav Melekhov, Surya Kalia, and Juho Kannala.
\newblock Camera relocalization by computing pairwise relative poses using
  convolutional neural network.
\newblock In {\em The IEEE International Conference on Computer Vision (ICCV)},
  Oct 2017.

\bibitem{RCDRTKArXiv}
Will Maddern, Geoffrey Pascoe, Matthew Gadd, Dan Barnes, Brian Yeomans, and
  Paul Newman.
\newblock Real-time kinematic ground truth for the oxford robotcar dataset.
\newblock 2020.

\bibitem{RobotCarDatasetIJRR}
Will Maddern, Geoff Pascoe, Chris Linegar, and Paul Newman.
\newblock {1 Year, 1000km: The Oxford RobotCar Dataset}.
\newblock volume~36, pages 3--15, 2017.

\bibitem{rotation_averaging}
F~Landis Markley, Yang Cheng, John~L Crassidis, and Yaakov Oshman.
\newblock Averaging quaternions.
\newblock {\em Journal of Guidance, Control, and Dynamics}, 30(4):1193--1197,
  2007.

\bibitem{nerfw}
Ricardo Martin-Brualla, Noha Radwan, Mehdi S.~M. Sajjadi, Jonathan~T. Barron,
  Alexey Dosovitskiy, and Daniel Duckworth.
\newblock {NeRF in the Wild: Neural Radiance Fields for Unconstrained Photo
  Collections}.
\newblock In {\em CVPR}, 2021.

\bibitem{scr_nn}
Daniela Massiceti, Alexander Krull, Eric Brachmann, Carsten Rother, and
  Philip~H.S. Torr.
\newblock Random forests versus neural networks — what's best for camera
  localization?
\newblock In {\em 2017 IEEE International Conference on Robotics and Automation
  (ICRA)}, pages 5118--5125, 2017.

\bibitem{nerf}
Ben Mildenhall, Pratul~P. Srinivasan, Matthew Tancik, Jonathan~T. Barron, Ravi
  Ramamoorthi, and Ren Ng.
\newblock Nerf: Representing scenes as neural radiance fields for view
  synthesis.
\newblock In {\em ECCV}, 2020.

\bibitem{coordinet}
Arthur Moreau, Nathan Piasco, Dzmitry Tsishkou, Bogdan Stanciulescu, and Arnaud
  de La~Fortelle.
\newblock Coordinet: uncertainty-aware pose regressor for reliable vehicle
  localization.
\newblock In {\em Proceedings of the IEEE/CVF Winter Conference on Applications
  of Computer Vision}, pages 2229--2238, 2022.

\bibitem{lens}
Arthur Moreau, Nathan Piasco, Dzmitry Tsishkou, Bogdan Stanciulescu, and Arnaud
  de La~Fortelle.
\newblock Lens: Localization enhanced by nerf synthesis.
\newblock In {\em Proceedings of the 5th Conference on Robot Learning}, volume
  164 of {\em Proceedings of Machine Learning Research}, pages 1347--1356.
  PMLR, 2022.

\bibitem{Park_2019_CVPR}
Jeong~Joon Park, Peter Florence, Julian Straub, Richard Newcombe, and Steven
  Lovegrove.
\newblock Deepsdf: Learning continuous signed distance functions for shape
  representation.
\newblock In {\em The IEEE Conference on Computer Vision and Pattern
  Recognition (CVPR)}, June 2019.

\bibitem{piasco2018survey}
Nathan Piasco, D{\'e}sir{\'e} Sidib{\'e}, C{\'e}dric Demonceaux, and
  Val{\'e}rie Gouet-Brunet.
\newblock A survey on visual-based localization: On the benefit of
  heterogeneous data.
\newblock {\em Pattern Recognition}, 74:90--109, 2018.

\bibitem{piasco_2019}
Nathan Piasco, Désiré Sidibé, Valérie Gouet-Brunet, and Cédric Demonceaux.
\newblock Learning scene geometry for visual localization in challenging
  conditions.
\newblock In {\em 2019 International Conference on Robotics and Automation
  (ICRA)}, pages 9094--9100, 2019.

\bibitem{roadmap}
Tong Qin, Yuxin Zheng, Tongqing Chen, Yilun Chen, and Qing Su.
\newblock A light-weight semantic map for visual localization towards
  autonomous driving.
\newblock In {\em 2021 IEEE International Conference on Robotics and Automation
  (ICRA)}, pages 11248--11254, 2021.

\bibitem{RARS19}
J. Revaud, J. Almazan, R.S. Rezende, and C.R. de Souza.
\newblock Learning with average precision: Training image retrieval with a
  listwise loss.
\newblock In {\em ICCV}, 2019.

\bibitem{saha2018improved}
Soham Saha, Girish Varma, and CV Jawahar.
\newblock Improved visual relocalization by discovering anchor points.
\newblock {\em arXiv preprint arXiv:1811.04370}, 2018.

\bibitem{sarlin2019coarse}
Paul-Edouard Sarlin, Cesar Cadena, Roland Siegwart, and Marcin Dymczyk.
\newblock From coarse to fine: Robust hierarchical localization at large scale.
\newblock In {\em CVPR}, 2019.

\bibitem{sarlin20superglue}
Paul-Edouard Sarlin, Daniel DeTone, Tomasz Malisiewicz, and Andrew Rabinovich.
\newblock {SuperGlue}: Learning feature matching with graph neural networks.
\newblock In {\em CVPR}, 2020.

\bibitem{sarlin2021pixloc}
Paul-Edouard Sarlin, Ajaykumar Unagar, Måns Larsson, Hugo Germain, Carl Toft,
  Viktor Larsson, Marc Pollefeys, Vincent Lepetit, Lars Hammarstrand, Fredrik
  Kahl, and Torsten Sattler.
\newblock {Back to the Feature}: Learning robust camera localization from
  pixels to pose.
\newblock In {\em CVPR}, 2021.

\bibitem{hyperpoints}
Torsten Sattler, Michal Havlena, Filip Radenovic, Konrad Schindler, and Marc
  Pollefeys.
\newblock Hyperpoints and fine vocabularies for large-scale location
  recognition.
\newblock In {\em 2015 IEEE International Conference on Computer Vision
  (ICCV)}, pages 2102--2110, 2015.

\bibitem{sattler2016large}
Torsten Sattler, Michal Havlena, Konrad Schindler, and Marc Pollefeys.
\newblock Large-scale location recognition and the geometric burstiness
  problem.
\newblock In {\em Proceedings of the IEEE conference on computer vision and
  pattern recognition}, pages 1582--1590, 2016.

\bibitem{Sattler2019}
Torsten Sattler, Qunjie Zhou, Marc Pollefeys, and Laura Leal-taix{\'e}.
\newblock {Understanding the Limitations of CNN-based Absolute Camera Pose
  Regression}.
\newblock In {\em CVPR}, pages 3302--3312, 2019.

\bibitem{cityscalelocation}
G. Schindler, M. Brown, and R. Szeliski.
\newblock City-scale location recognition.
\newblock In {\em Proceedings of the International Conference on Computer
  Vision and Pattern Recognition (CVPR07)}, Minneapolis, June 2007.

\bibitem{shavit2021learning}
Yoli Shavit, Ron Ferens, and Yosi Keller.
\newblock Learning multi-scene absolute pose regression with transformers.
\newblock In {\em Proceedings of the IEEE/CVF International Conference on
  Computer Vision}, pages 2733--2742, 2021.

\bibitem{TransPoseNet}
Yoli Shavit, Ron Ferens, and Yosi Keller.
\newblock Paying attention to activation maps in camera pose regression.
\newblock In {\em arxiv preprint, arxiv:2103.11477}, 2021.

\bibitem{forest_SCR}
Jamie Shotton, Ben Glocker, Christopher Zach, Shahram Izadi, Antonio Criminisi,
  and Andrew Fitzgibbon.
\newblock Scene coordinate regression forests for camera relocalization in
  rgb-d images.
\newblock In {\em Proc. Computer Vision and Pattern Recognition (CVPR)}. IEEE,
  June 2013.

\bibitem{sitzmann2019siren}
Vincent Sitzmann, Julien~N.P. Martel, Alexander~W. Bergman, David~B. Lindell,
  and Gordon Wetzstein.
\newblock Implicit neural representations with periodic activation functions.
\newblock In {\em Proc. NeurIPS}, 2020.

\bibitem{sitzmann2019srns}
Vincent Sitzmann, Michael Zollh{\"o}fer, and Gordon Wetzstein.
\newblock Scene representation networks: Continuous 3d-structure-aware neural
  scene representations.
\newblock In {\em Advances in Neural Information Processing Systems}, 2019.

\bibitem{song2022dalg}
Yuxin Song, Ruolin Zhu, Min Yang, and Dongliang He.
\newblock Dalg: Deep attentive local and global modeling for image retrieval.
\newblock {\em arXiv preprint arXiv:2207.00287}, 2022.

\bibitem{Sucar:etal:ICCV2021}
Edgar Sucar, Shikun Liu, Joseph Ortiz, and Andrew Davison.
\newblock {iMAP}: Implicit mapping and positioning in real-time.
\newblock In {\em Proceedings of the International Conference on Computer
  Vision ({ICCV})}, 2021.

\bibitem{tancik2020fourier}
Matthew Tancik, Pratul Srinivasan, Ben Mildenhall, Sara Fridovich-Keil, Nithin
  Raghavan, Utkarsh Singhal, Ravi Ramamoorthi, Jonathan Barron, and Ren Ng.
\newblock Fourier features let networks learn high frequency functions in low
  dimensional domains.
\newblock {\em Advances in Neural Information Processing Systems},
  33:7537--7547, 2020.

\bibitem{torii201524}
Akihiko Torii, Relja Arandjelovic, Josef Sivic, Masatoshi Okutomi, and Tomas
  Pajdla.
\newblock 24/7 place recognition by view synthesis.
\newblock In {\em Proceedings of the IEEE conference on computer vision and
  pattern recognition}, pages 1808--1817, 2015.

\bibitem{torii2011visual}
Akihiko Torii, Josef Sivic, and Tomas Pajdla.
\newblock Visual localization by linear combination of image descriptors.
\newblock In {\em 2011 IEEE International Conference on Computer Vision
  Workshops (ICCV Workshops)}, pages 102--109. IEEE, 2011.

\bibitem{von2020gn}
Lukas Von~Stumberg, Patrick Wenzel, Qadeer Khan, and Daniel Cremers.
\newblock Gn-net: The gauss-newton loss for multi-weather relocalization.
\newblock {\em IEEE Robotics and Automation Letters}, 5(2):890--897, 2020.

\bibitem{AtLoc}
Bing Wang, Changhao Chen, Chris Xiaoxuan~Lu, Peijun Zhao, Niki Trigoni, and
  Andrew Markham.
\newblock Atloc: Attention guided camera localization.
\newblock In {\em Proceedings of the AAAI Conference on Artificial
  Intelligence}, volume~34, pages 10393--10401, 2020.

\bibitem{wenzel2020fourseasons}
P. Wenzel, R. Wang, N. Yang, Q. Cheng, Q. Khan, L. von Stumberg, N. Zeller, and
  D. Cremers.
\newblock {4Seasons}: A cross-season dataset for multi-weather {SLAM} in
  autonomous driving.
\newblock In {\em Proceedings of the German Conference on Pattern Recognition
  ({GCPR})}, 2020.

\bibitem{xie2021neuralfield}
Yiheng Xie, Towaki Takikawa, Shunsuke Saito, Or Litany, Shiqin Yan, Numair
  Khan, Federico Tombari, James Tompkin, Vincent Sitzmann, and Srinath Sridhar.
\newblock Neural fields in visual computing and beyond, 2021.

\bibitem{graph_neural_network}
F. {Xue}, X. {Wu}, S. {Cai}, and J. {Wang}.
\newblock Learning multi-view camera relocalization with graph neural networks.
\newblock In {\em 2020 IEEE/CVF Conference on Computer Vision and Pattern
  Recognition (CVPR)}, pages 11372--11381, 2020.

\bibitem{Zhou_2019_CVPR}
Yi Zhou, Connelly Barnes, Lu Jingwan, Yang Jimei, and Li Hao.
\newblock On the continuity of rotation representations in neural networks.
\newblock In {\em The IEEE Conference on Computer Vision and Pattern
  Recognition (CVPR)}, June 2019.

\bibitem{DA4AD_2020_ECCV}
Yao Zhou, Guowei Wan, Shenhua Hou, Li Yu, Gang Wang, Xiaofei Rui, and Shiyu
  Song.
\newblock Da4ad: End-to-end deep attention-based visual localization for
  autonomous driving.
\newblock In {\em Proceedings of the European Conference on Computer Vision
  (ECCV)}, 2020.

\bibitem{zhu_cvpr}
Y. Zhu, R. Gao, S. Huang, S. Zhu, and Y. Wu.
\newblock Learning neural representation of camera pose with matrix
  representation of pose shift via view synthesis.
\newblock In {\em 2021 IEEE/CVF Conference on Computer Vision and Pattern
  Recognition (CVPR)}, pages 9954--9963, Los Alamitos, CA, USA, jun 2021. IEEE
  Computer Society.

\end{thebibliography}
